\documentclass[10pt,twocolumn]{article}

\usepackage{cvpr}
\usepackage{times}
\usepackage{epsfig}
\usepackage{graphicx}
\usepackage{amsmath}
\usepackage{amssymb}
\usepackage{soul}
\usepackage{multirow}

\usepackage[pagebackref=true,breaklinks=true,colorlinks,bookmarks=false]{hyperref}

\cvprfinalcopy % *** Uncomment this line for the final submission

\def\cvprPaperID{****} % *** Enter the CVPR Paper ID here

\ifcvprfinal\fi
\begin{document}

%%%%%%%%% TITLE
\title{The Hateful Memes Challenge Next Move}

\author{Weijun Jin\\
Georgia Institute of Technology\\
North Ave NW, Atlanta, GA 30332\\
{\tt\small wjin62@gatech.edu}
\and
Lance Wilhelm\\
Georgia Institute of Technology\\
North Ave NW, Atlanta, GA 30332\\
{\tt\small lwihelm7@gatech.edu}
}

\maketitle
%\thispagestyle{empty}

%%%%%%%%% ABSTRACT
\begin{abstract}

State-of-the-art image and text classification models, such as Convolutional Neural Networks and Transformers, have long been able to classify their respective unimodal reasoning satisfactorily with accuracy close to or exceeding human accuracy. However, images embedded with text, such as hateful memes, are hard to classify using unimodal reasoning when difficult examples, such as benign confounders, are incorporated into the data set. We attempt to generate more labeled memes in addition to the Hateful Memes data set from Facebook AI, based on the framework of a winning team from the Hateful Meme Challenge. To increase the number of labeled memes, we explore semi-supervised learning using pseudo-labels for newly introduced, unlabeled memes gathered from the Memotion Dataset 7K. We find that the semi-supervised learning task on unlabeled data required human intervention and filtering and that adding a limited amount of new data yields no extra classification performance.
\end{abstract}

%%%%%%%%% BODY TEXT
\section{Introduction/Background/Motivation}

\begin{figure}
\centering
\includegraphics[width=0.8\linewidth]{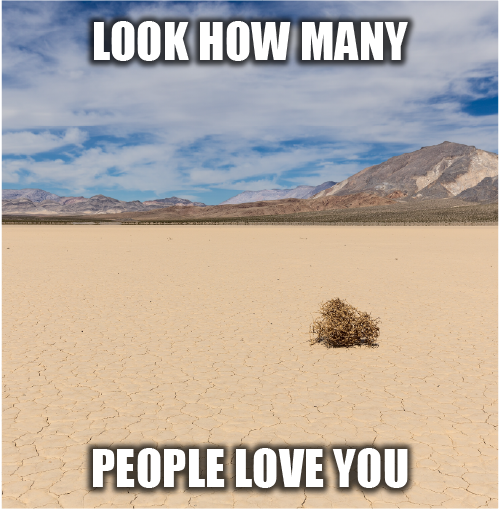}
\caption{An example of a mean meme. Because hateful memes are generally sexist, racist, etc., we chose not to show any hateful memes used to train the model in this paper. Nonetheless, this image depicts the challenge that faces multimodal meme classification whereby the text and the photo confound one another.}
\label{fig:meanmeme}
\end{figure}

Many real-world problems are multimodal in nature, such as how humans interact with the environment around them based on various real-world objects. To measure the role truly multimodal reasoning plays in understanding and solving many current tasks, Kiela~\etal \cite{HatefulMemesChallenge} at Facebook AI created a dataset and a challenge consisting of over 12,000 hateful and non-hateful memes from which a multimodal model could be trained to classify the hatefulness of a meme. An example of a meme similar to the hateful memes that are used to train our model can be found in figure \ref{fig:meanmeme}.

To illustrate the power of multimodal vs. unimodal reasoning, they evaluated the performance of unimodal models, multimodal models with unimodally pretrained, and multimodal models with multimodally pretrained based on this dataset. Although, as expected, the multimodal models outperform the unimodal models, all models, both unimodal and multimodal, struggle to achieve a similar hatefulness classification accuracy as the human's, 84.70\%. In their classifications tests on the Hateful Memes dataset, Kiela~\etal \cite{HatefulMemesChallenge} tried models such as RESNET152~\cite{Resnet}, which is a unimodal model used for image feature extraction, BERT~\cite{BERT} which is a unimodal model used for text classification, Visual BERT~\cite{visualBERT} which is a multimodal model unimodally pretrained, and Visual BERT COCO which is the same multimodal model but pretrained on the multimodal objective with COCO dataset~\cite{COCO}. The model that produced the closest accuracy to the human was the Visual BERT COCO model~\cite{visualBERT, COCO} which achieved a test accuracy of $69.47\pm2.06$ and test AUROC of $75.44\pm1.86$. 

Our goal is to improve the performance of Visual BERT COCO, the best baseline model in Kiela~\etal\cite{HatefulMemesChallenge}, and achieve results closer to or better than the human accuracy, along with hyper-parameter tuning. Considering deep learning is data-driven, we decided to apply semi-supervised learning to re-label data from Memotion Dataset 7K to expand the training dataset to improve performance. Since Visual BERT COCO and other SoTA models already exist in Facebook Research's repository, we can use their code as a starting point.

In the past few years, there has been a surge of interest in multimodal problems, from image captioning to visual question answering (VQA), according to Kiela~\etal\cite{HatefulMemesChallenge}. Sleeman \etal state that "there has been an increased focus on combining data from multiple modalities to further improve machine learning based classification models \cite{MultimodalLandscape}." The complicating factor regarding multimodal data comes when the representations of the data are starkly different from one another. Efforts must be taken to process the data and then combine them so that a model can perform its intended task. Multimodal learning can be divided into five challenges: representation, translation, alignment, fusion, and co-learning. With regard to hateful meme classification, we argue that the SoTA multimodal models successfully address all five of the challenges but struggle when it comes to understanding the complexity of confounding memes whereby the text and the image may contradict one another. However, not all confounding memes may be hateful, highlighting the importance of having a large and thorough dataset from which to train the algorithm. 

An important observation made by Kiela~\etal \cite{HatefulMemesChallenge} when analyzing the results from their baseline models is the importance of pretraining on model performance. One can observe that there is little improvement between models that were unimodally vs. multimodally pretrained. This corroborates the findings from Singh~\etal \cite{Pretraining}, showing the importance of using in-domain rather than out-of-domain pretraining. 

There is a clear use case for the classification of hate speech within social media as it has become so prevalent given the ubiquitous access that most of the world has to the internet. Because of that prevalence, social media companies are placed in a position of power to control the content within their platforms, which can infringe upon free speech rights within countries that protect such rights. As highlighted by Ring in their 2013 dissertation on hate speech in social media \cite{HateSpeechReview}, social media company content removal teams are solely responsible for detecting and removing hateful content within their platforms. Having a model that could confidently remove hateful memes immediately without human interaction and at a human level of accuracy could reduce the amount of overhead required to reduce hate speech and limit the spread of hate speech by drastically reducing the time it is exposed on the internet. There is evidence that social media can impact large portions of the global population and influence such things as elections \cite{ElectionSocialMedia, ElectionSocialMedia2, ElectionSocialMedia3}, and civic engagement \cite{CivicEngagement}. Social media's power to strongly influence global society underlines the importance of proper control.

The Hateful Memes dataset provided by Kiela~\etal \cite{HatefulMemesChallenge} at Facebook AI is the foundational dataset used to train any classification of hateful memes, and it appears to be one of if not the most thorough datasets in this domain publicly available. It was designed to emphasize the importance of multimodal reasoning with multimodally pretraining. Its purpose is to further hate speech detection and further multimodal model development. The dataset comprises five different types of memes: multimodal hate, unimodal hate, benign image, and benign text confounders, and finally, random not-hateful examples. The dataset is split into a training set and then into a dev and test set from 5\% and 10\% of the data, respectively. The dev and test set are fully balanced and are comprised of memes using the following percentages: 40\% multimodal hate, 10\% unimodal hate, 20\% benign text confounder, 20\% benign image confounder, and 10\% random non-hateful.

Each image in the dataset has associated metadata with it that gives the text contained within the meme and a label generated from human annotation and consensus of whether or not a meme is hateful. Having the text already within the metadata is nice as it allows us not to worry about text recognition within images. However, it is worth noting that the overall winner of the Hateful Memes Challenge, Zhu~\etal\cite{HatefulMemesWinner1}, relied upon optical character recognition (OCR) to detect the text within an image and remove it to provide the image classification portion of their model with a clean image to work with.

We also used the Memotion Dataset 7k from Sharma~\etal \cite{MemotionDataset} to add memes to the dataset from which to fine-tune the model. This dataset is a collection of 7000 memes collected by the authors. The accompanying metadata includes the text contained within the meme, a hatefulness label, a sentiment label, and a humor label. It is unclear how those labels were decided for each meme in the dataset, and Velioglu ~\etal\cite{HatefulMemesWinner3} found that the dataset is poorly labeled. Therefore, one needs to label the dataset. They went through the dataset and cherry-picked only 328 memes that were suitable for the Hateful Memes Challenge and added them to the training dataset. Considering the scarcity of publicly available memes datasets, we decide to apply semi-supervised learning to more than 6000 remaining memes to further expand the training dataset.

%------------------------------------------------------------------------
\section{Approach}

\begin{figure}
\includegraphics[width=1.0\linewidth]{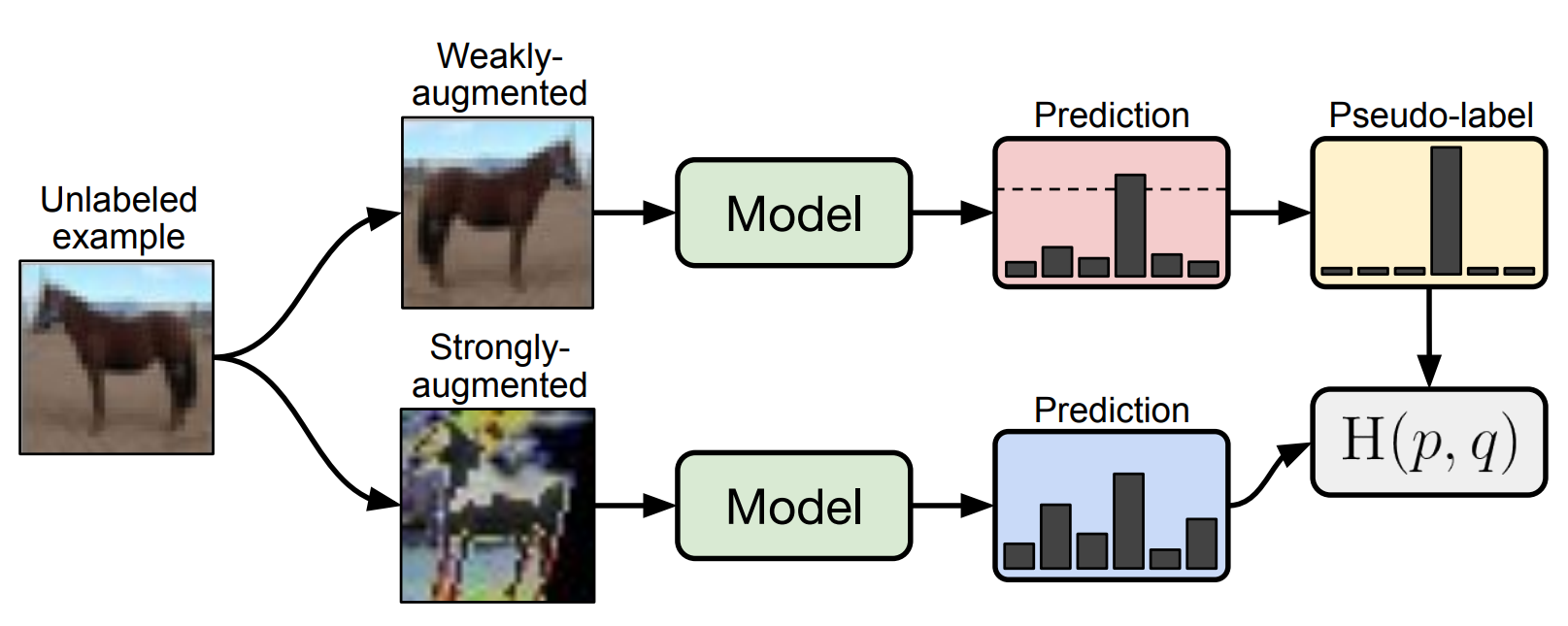}
\caption{FixMatch diagram}
\label{fig:fixmatch}
\end{figure}

Our approach is to further improve upon the success of the winners of the Hateful Memes Challenge. To do so we would focus on expanding the dataset by utilizing semi-supervised learning, specifically in a very similar architecture as the FixMatch algorithm introduced by Sohn~\etal\cite{FixMatch}. A diagram of the FixMatch architecture can be found in Figure \ref{fig:fixmatch}. 

We attempted to leverage the success of Velioglu and Rose \cite{HateDetectronPaper} in their implementation and fine-tuning of the Visual BERT model and expanded dataset. In their paper, they note how limited the Hateful Memes dataset is with only 10,000 images when compared to the large datasets used for training image and object classification models such as VisualGenome \cite{VisualGenome}, which contains 108,077 images, COCO \cite{COCO} which contains over 330K images, and Conceptual Captions \cite{ConceptualCaptions} which contains over 3.3M images. In their winning method, Velioglu and Rose \cite{HateDetectronPaper} attempted to add more data into the labeled training pool of memes from the unused data in dev\_seen as well as the Memotion Dataset 7K \cite{MemotionDataset}, but because of poor labeling, they only added 328 additional memes. Because of the relative lack of data for training the model, we attempted to add additional training data using a semi-supervised method which should create more stable learning and better results.

First, we needed to set up the environment and the model in the same way that Team HateDetectron did in their successful implementation found at their GitHub repository \cite{HateDetectronGithub}. This meant the installation of Facebook AI's Multimodal Framework (MMF) package \cite{MMF}, followed by downloading and ingesting the 12,140 memes from the Hateful Memes dataset into the framework. Once the data had been loaded into the framework, MMF has built-in functions described in its documentation that allow the loading, running, and training of preexisting models or the creation of brand new models that can be trained on the Hateful Memes dataset. 

\begin{figure}
\includegraphics[width=1.0\linewidth]{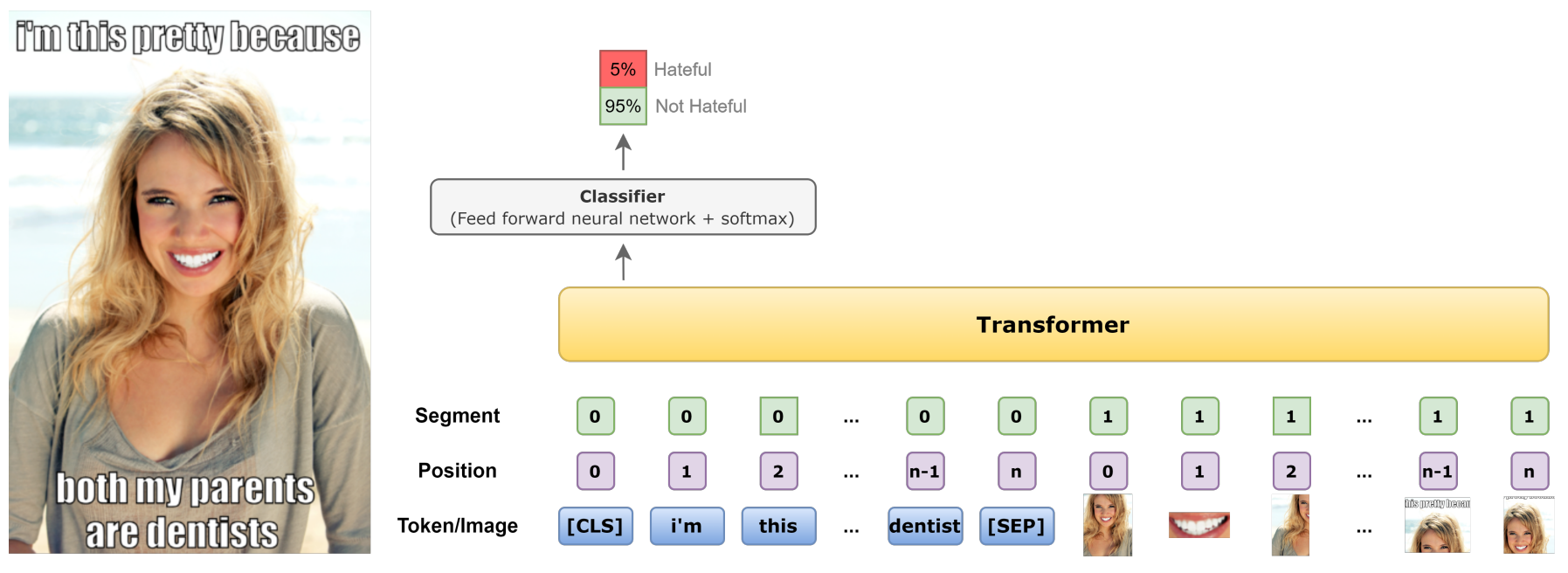}
\caption{An example of the Visual BERT process}
\label{fig:visualbert}
\end{figure}

Next, we downloaded the Memotion 7K dataset \cite{MemotionDataset} and loaded them into the same image folder in which the Hateful Memes dataset was loaded. What is important about our implementation at this point is that we decided to include all of the memes from the Memotion 7K dataset as opposed to only including the 328 labeled memes by Velioglu and Rose \cite{HateDetectronPaper}. This brought the total number of memes in the dataset up to 19,132. From here, the most important part of the improved model's performance comes in the form of extracting features from each of the images in the dataset using a "ResNeXT-152 based Mask-RCNN model trained on VisualGenome...these visual embeddings are then projected in the textual embedding space before passing them through the transformer layers" \cite{HateDetectronPaper} of the Visual BERT model. Figure \ref{fig:visualbert} shows an example of this. 

Per observation, the feature extraction process ignores tens of memes from the Memotion 7K dataset due to irregular size. As the result, 19,091 memes have features extracted rather than 19,132. Next, we created training metadata comprising 8,928 reliably labeled memes with 8,500 memes from the Hateful Meme training dataset, 100 memes from the hateful meme Dev\_seen dataset that are not in the 'dev\_unseen dataset, and 328 memes from Memotion 7K dataset. These features, training metadata, and hyperparameters are passed to the Visual BERT model pretrained on COCO for training. Once the model is trained, we can obtain the resulting \verb|*.ckpt| files, which contain the learned parameters for our fine-tuned models. 

Next, we created new metadata based on the rest of the Memotion 7K dataset, excluding 328 memes for semi-supervised learning to assign the pseudo labels. The initial number of memes in the forecast metadata is 6,664. Based on initial screening, we found some memes text in the test metadata containing "none" or an empty string. This causes the tokenizing BERT to fail in splitting the string when fed into the Visual BERT model for prediction. Consequently, we removed these irregular texts from the forecast metadata, with the number of memes reduced to 6,623.

Our model differs from that of Velioglu and Rose's model because we will generate pseudo-labels for the 6,623 remaining unlabeled memes from the Memotion dataset and then perform a FixMatch-like semi-supervised learning process with them. After pseudo labels are assigned in the forecast output metadata, we set a threshold greater than or equal to 0.995 for positive pseudo labels or a threshold less than or equal to 0.005 for negative pseudo labels upon the probability column to filter the pseudo label memes. The resulting number of pseudo-labeled memes is 1,534 after filtering. At the moment we trusted the reliability of assigned pseudo labels with such extreme confidence.

\begin{figure}
\centering
\includegraphics[width=0.8\linewidth]{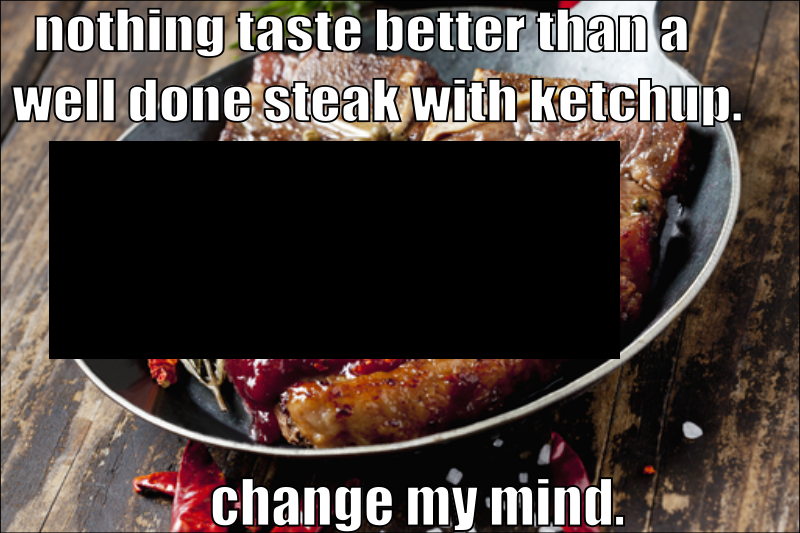}
\caption{An example of a meme with a random cutout transformation applied.}
\label{fig:cutoutmeme}
\end{figure}

Next, based on the FixMatch-like semi-supervised learning process, we applied the image transformation, such as cutout, to 1,534 pseudo-labeled memes to create strongly augmented memes to replace original memes. An example of a meme with a cutout applied can be found in Figure \ref{fig:cutoutmeme}. We then concatenated the 1,534 memes with their original training metadata and pseudo labels, which increased the number of memes in the new training metadata to 10,462. We retrained the Visual BERT model with the new training metadata and the same hyperparameters set. 

Lastly, we evaluated the model performance upon existing hateful memes unseen\_testing metadata in terms of accuracy and AUROC to measure the impact of semi-supervised learning. We iterated and tuned the above process to continue adding newly assigned pseudo-labeled memes into old training metadata or cherry-picking already assigned pseudo-labeled memes in the old training metadata based on in-domain knowledge until we believed in no room for model performance improvement based on the analysis of data assumption for pseudo labeled memes.

\begin{figure}
\centering
\includegraphics[width=0.8\linewidth]{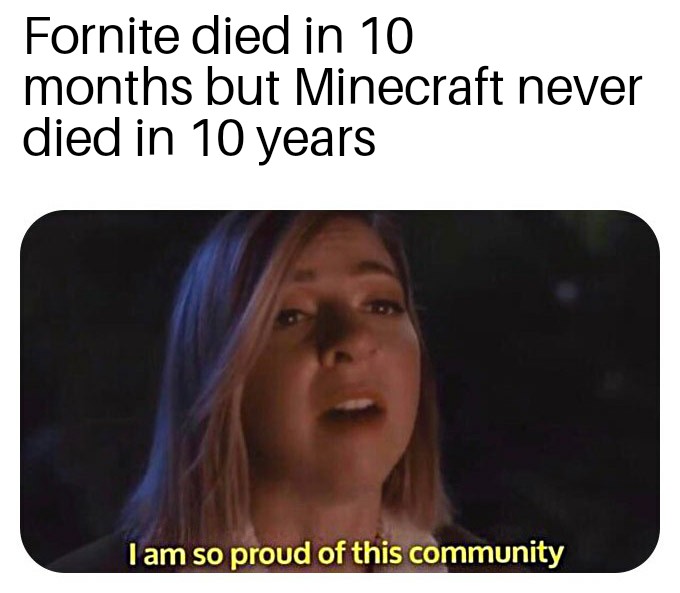}
\caption{An example of a meme misclassified as hateful with a confidence greater than 99.5\% by the initial model. This meme displays positive affection toward the community of the video game Minecraft for staying active for longer than other video games. The image also shows a woman displaying positive affection, which does not create any confusion between the text and the image. It is possible that the model is overly sensitive to words such as "died."}
\label{fig:misclassifiedmeme}
\end{figure}

Initially, we assumed the extreme threshold such as 0.995 and 0.005 around two tails, can give a certain degree of reliability for assigned pseudo labels. Still, we found some memes pseudo-labeled as hateful actually should not be based on our sanity check. An example of a meme misclassified as hateful is given in figure \ref{fig:misclassifiedmeme}. 

We found with these model-selected pseudo-labeled memes incorporated in the training metadata, the model performance metrics of accuracy and AUROC on the test dataset were not improved. This result contradicts our initial expectation. We acknowledged using semi-supervised learning without human review and modification is infeasible. Consequently, we reviewed 822 pseudo-labeled hateful memes in the training metadata and found only 282 might be valid based on our analysis. Considering training metadata imbalanced with less proportion for hateful memes, we decided to incorporate 282 cherry-picked pseudo-labeled hateful memes into training metadata. We retrained the model and found that the model performance metrics were still not effectively improved. 

Our approach leveraged preexisting notebooks from Facebook AI and one of the winning teams, HateDetectron \cite{HateDetectronGithub}. Due to the competition occurring 2 years ago, the internal structure of the downloaded hateful meme dataset zip file is no longer fit for the original unzipped code, so we modified the internal structure of the zip file. In addition, we modified the code for memes merger and training metadata concatenation for a different data source to streamline the semi-supervised learning process and added data mining tools to facilitate memes review and filtering. We figured out how to randomly create the augmented images using the \verb|torchvision| package, which allowed us to keep all image manipulations within Python. We attempted multiple combinations of hyperparameters fine-tuning.

\section{Experiments and Results}

The goal is to enhance the performance metrics of the best baseline model, Visual Bert COCO in Kiela~\etal\cite{HatefulMemesChallenge} to approach the human's test accuracy of 84.7\% on hateful memes classification. We leveraged Memotion Dataset 7K \cite{MemotionDataset} to increment the original training metadata from Facebook AI via semi-supervised learning. There are 3 experiments upon 3 stages of incremental training metadata and the data characteristic analysis. The major performance metrics are accuracy and AUROC. AUROC is the area under the receiver operating characteristic curve based on the integration of the product of true positive rate and inverted false positive rate between 0 and 1. 

Firstly we fine-tuned the Visual Bert COCO as the baseline based on original training metadata from Facebook AI, along with 2 sets of hyperparameters. We then set the loss function as cross-entropy in the classification layer and the optimization algorithm as Adam, both of which are standard in deep learning. Next, we set the visual embedding dimension as 2048 because the ResNeXT-152-based Mask-RCNN model extracts 100 boxes of 2048D region-based image features from an fc6 layer. We then conducted a grid search to identify 2 sets of hyperparameters that can optimize the model performance. We decided to implement a learning rate decay schedule which allows the model to have big steps using a high learning rate to explore the optimal solution at the start of training, then, after a while, the model decayed the learning rate to refine the search. Empirically warm-up steps can not exceed half of the total steps because too long of a warm-up will bump up across the valleys of the objective function. We attempted different decay functions, such as linear vs. non-linear, to test the efficacy. Theoretically, linear decay is usually along with longer warm-up steps in table \ref{tab:hyperparameters} vs. non-linear decay because linear decay is faster. We attempted different learning rates and different learning schedules to test the efficiency. We also attempted different batch sizes to check the impact of randomness on the optimization. Empirically small batch sizes can add more noise in the training and lead to more randomness in the search for optimal solutions. We attempted a different number of training steps to check the sufficiency for converging to an optimal solution. Empirically, incremental epoch sizes can help a calibrated model reach the optimal solution. Eventually, we derived 2 sets of best candidates for hyperparameters in table \ref{tab:hyperparameters}.

\begin{table}
\centering
\begin{tabular}{l|cc}
\hline
 & \multicolumn{2}{c}{\textbf{Model}} \\
\textbf{Hyperparameters} & 1 & 2 \\ \hline
Epoch & 3000 & 3500 \\ \hline
Learning Rate & 0.3 & 0.6 \\ \hline
Learning Rate Decay & warmup\_linear & warmup\_cosine \\ \hline
Warm Up & TRUE & TRUE \\ \hline
Warm Up Steps & 2000 & 500 \\ \hline
Batch Size & 32 & 80 \\ \hline
Optimizer Learning Rate & 5e-5 & 5e-5 \\ \hline
\end{tabular}
\caption{Best model hyperparmeters}
\label{tab:hyperparameters}
\end{table}

At stage 1, we incremented training metadata by 328 memes from Memotion Dataset 7K \cite{MemotionDataset} manually picked by Velioglu ~\etal\cite{HatefulMemesWinner3}. At stage 2, we incremented training metadata by 1534 via semi-supervised learning using strong augmentation. Lastly, at stage 3, we manually re-labeled hateful pseudo labels, which were part of the 1534 pseudo labels selected for increasing the training dataset at stage 2. We found the number of actual hateful memes might only be 282 based on human annotations. We decided to replace the 1534 pseudo-labeled memes with the 282 manually labeled memes.

\begin{figure}
\centering
\includegraphics[width=\linewidth]{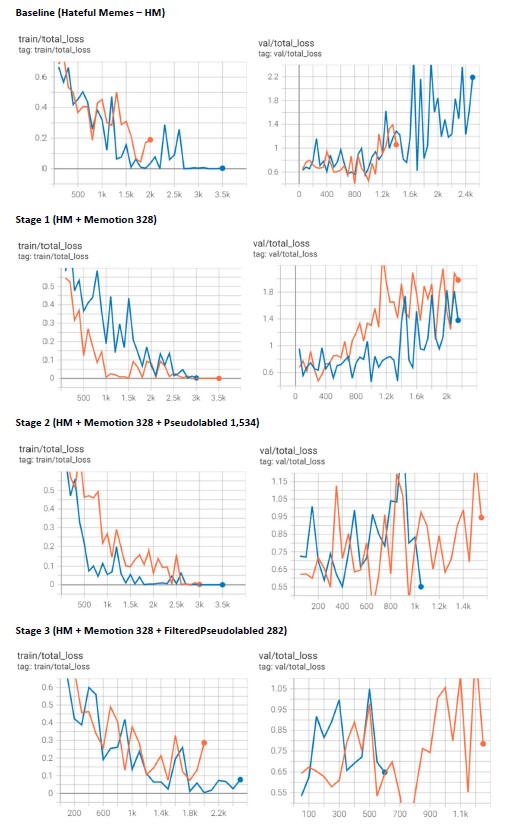}
\caption{loss function}
\label{fig:loss function}
\end{figure}

Per the loss function for baseline and stages 1 - 3 in Figure \ref{fig:loss function}, we found the model 1 - set 1 hyperparameters (orange) and the model 2 - set 2 hyperparameters (blue) have overfitting because the training loss is lower than validation loss. The default dropout rate is 10\%, but we didn't override it with a higher rate considering testing performance metrics are better than validation in Table \ref{tab:resultstable}.

The Visual BERT model is particularly attuned to the structure of our problem because we are trying to fit a model to a contextual sequence of text and an image. The Visual BERT model builds on top of the groundbreaking natural language processing model, BERT \cite{BERT}, which pioneered using bi-directional self-attention layers to understand the context from every part of a text sequence at once. Similarly, Visual BERT utilizes an RCNN model to provide the BERT model image tokens alongside the text tokens. This should, in theory, allow the model to extract contextual features between the text sequence and the text and image sequence, making it truly multimodal.

Per performance metrics found in Table \ref{tab:resultstable}, as we can see, no matter how we incremented and manually modified pseudo labels in the training metadata, we can not significantly improve the model performance. 

\begin{table*}
\centering
\begin{tabular}{l|l|cc|cc}
\hline
 &  & \multicolumn{2}{c|}{\textbf{Validation}} & \multicolumn{2}{c}{\textbf{Test}} \\ 
\textbf{Dataset} & \textbf{Model} & Acc. & AUROC & Acc. & AUROC \\ \hline
\multirow{2}{*}{Hateful Memes (HM) (Baseline)} & 1 & 68.33 & 73.59 & 70.30 & 75.23 \\ \cline{2-6} & 2 & 70.37 & 72.91 & 72.60 & 76.84 \\ \hline
\multirow{2}{*}{HM + Memotion 328} & 1 & 71.30 & 74.07 & 72.35 & 76.87 \\ \cline{2-6} & 2 & 74.26 & 74.57 & 73.10 & 78.45 \\ \hline
\multirow{2}{*}{HM + Memotion 328 + Pseudo-labeled 1,534} & 1 & 69.07 & 75.00 & 69.80 & 75.88 \\ \cline{2-6} & 2 & 70.00 & 75.28 & 72.30 & 77.83 \\ \hline
\multirow{2}{*}{HM + Memotion 328 + Filtered Pseudo-labeled 282} & 1 & 70.56 & 75.14 & 71.90 & 76.55 \\ \cline{2-6} & 2 & 68.33 & 73.50 & 69.85 & 74.28 \\ \hline
\end{tabular}
\caption{Model performance}
\label{tab:resultstable}
\end{table*}

%-------------------------------------------------------------------------

Based on the comparison of hateful meme characteristics (qualitative analysis) between testing and pseudo-labeled training metadata, we found a gap in the data assumption, indicating pseudo-labeled memes usually are unimodal. In contrast, those in testing from the Facebook Hateful Memes dataset are mostly multimodal. This is likely due to the careful construction of the original Hateful Memes dataset by Kiela~\etal \cite{HatefulMemesChallenge} with special consideration to the proportions of types of memes.

Visual BERT is in the family of transformer models. The special attribute of the transformer is the self-attention layer, usually along with multi-head attention. Because the self-attention layer comprises a query vector, key vector, and value vector, the important learnable parameters are key, value, and query matrices for the affine transformation of input variables. Other important learnable parameters include weights of hidden layers, feed-forward layers, and fully connected classification layers.

\section{Future Work}

Other approaches to this problem that could be explored in the future include implementing new classification layers for measuring model confidence, analyzing pretraining on datasets such as COCO to improve transfer learning, using external sources to add additional information about the content of images, and using better text encoders to replace the BERT text encoder. Most of these tasks involve a change in the model architecture, which is how most of the winning teams from the Hateful Memes Challenge achieved increased performance. While changing the model architecture may achieve greater performance, it is more complicated than exploring dataset modifications. Changes to model architecture require an intimate knowledge of the current SoTA models to seek out areas for improvement within the models. Kiela~\etal \cite{HatefulMemesChallenge} showed in their initial report that achieving model performance is not as simple as fusing two successful models. Greater consideration must be given to the underlying characteristic of a model to achieve greater performance.

%-------------------------------------------------------------------------
\section{Conclusion}

In our work, we explored how we could improve the performance in the Hateful Memes Challenge by introducing more data into the training set using a semi-supervised learning technique. Classifying hateful memes accurately could help limit the spread of hate speech on the internet. Detecting whether or not a meme is hateful is a tough task because of the multimodal nature of memes. Previous SoTA models have limited success when compared to a human. Our work, which leveraged the successful work of Team HateDetectron from the Hateful Memes Challenge, showed that semi-supervised learning for hateful meme detection is tough to accomplish without human interaction and filtering. We also showed that adding different structured data to the training dataset had no appreciable effect on the performance of the classification model. We hope this work can help guide success in future works to help reduce the prevalence of hate speech on the internet. 

%-------------------------------------------------------------------------
% \section{Work Division}

% The distribution of work for this assignment can be found in Table \ref{tab:contributions}. It is worth noting that we started this class with 4 members on our team before two of them dropped off our team without notice (and possibly out of the class) after the project proposal. This caused us to pivot our project to something more reasonable for 2 members in the short time that we had between this project and the class readings, quizzes, and assignments.

% \begin{table*}
% \begin{center}
% \begin{tabular}{|l|c|p{8cm}|}
% \hline
% Student Name & Contributed Aspects & Details \\
% \hline\hline
% Weijun Jin & Implementation and Analysis & Trained model and extracted features. Produced the predictions on unlabeled data. Test predictions. Report authoring.\\
% Lance Wilhelm & Implementation and Analysis & Trained model. Produced a new augmented dataset from confident pseudo-labels. Test predictions. Report authoring.\\
% \hline
% \end{tabular}
% \end{center}
% \caption{Contributions of team members.}
% \label{tab:contributions}
% \end{table*}

\newpage

{\small
\bibliographystyle{ieee_fullname}
\bibliography{egbib}
}

\end{document}